\documentclass[letterpaper, 10 pt, conference]{ieeeconf}  % Comment this line out if you need a4paper
\usepackage{amsmath,amsfonts}
\usepackage{algorithmic}
\usepackage{algorithm}
\usepackage{array}
\usepackage{textcomp}
\usepackage{url}
\usepackage{verbatim}
\usepackage{graphicx}
\usepackage{cite}
\usepackage{stfloats}
\usepackage{booktabs}
\usepackage{color}
\usepackage[dvipsnames]{xcolor}
\usepackage{dsfont}
\usepackage{colortbl}
\usepackage{bm}
\usepackage{hyperref}
\hypersetup{
    colorlinks = true,
    linkcolor = blue,
    citecolor = blue,
    urlcolor = blue
}
\usepackage{multirow}
\usepackage{soul}
\usepackage{ragged2e}

\usepackage{multicol}
\usepackage{makecell}
\usepackage{booktabs}
\usepackage{amssymb}
\usepackage{bbding}
\usepackage{pifont}
\usepackage{wasysym}
\usepackage{subcaption}
\usepackage{siunitx}
\definecolor{my_blue}{rgb}{0, 0.4470, 0.7410}
\definecolor{my_yellow}{rgb}{0.9290, 0.6940, 0.1250}
\definecolor{my_purple}{rgb}{0.4940, 0.1840, 0.5560}
\definecolor{my_green}{rgb}{0.4660, 0.6740, 0.1880}
\definecolor{my_red}{rgb}{0.6350, 0.0780, 0.1840}
\definecolor{my_black}{rgb}{0.25, 0.25, 0.25}
\definecolor{my_turquoise}{rgb}{0.000, 0.710, 0.694}
\definecolor{my_pink}{rgb}{0.910, 0.263, 0.576}

\IEEEoverridecommandlockouts                              
\title{ \Large\bf
Learning-based Trajectory Tracking for Bird-inspired Flapping-Wing Robots\\

\author{Jiaze Cai*,
        Vishnu Sangli*, 
        Mintae Kim
        and 
        Koushil Sreenath}
\thanks{* Equal Contribution. The authors are with the Hybrid Robotics Lab, University of California, Berkeley, CA 94720, United States.}
}

\begin{document}
\maketitle
\thispagestyle{empty}
\pagestyle{empty}

\begin{abstract}
Bird-sized flapping-wing robots offer significant potential for agile flight in complex environments, but achieving agile and robust trajectory tracking remains a challenge due to the complex aerodynamics and highly nonlinear dynamics inherent in flapping-wing flight. In this work, a learning-based control approach is introduced to unlock the versatility and adaptiveness of flapping-wing flight. We propose a model-free reinforcement learning (RL)-based framework for a high degree-of-freedom (DoF) bird-inspired flapping-wing robot that allows for multimodal flight and agile trajectory tracking. Stability analysis was performed on the closed-loop system comprising of the flapping-wing system and the RL policy. Additionally, simulation results demonstrate that the RL-based controller can successfully learn complex wing trajectory patterns, achieve stable flight, switch between flight modes spontaneously, and track different trajectories under various aerodynamic conditions. 
\end{abstract}
  
\section{Introduction}
Flapping-wing Micro Aerial Vehicles (FMAVs) offer tremendous potential to achieve efficient and agile flight in complex, cluttered environments by mimicking the capabilities of their biological counterparts. These vehicles, capable of simultaneously generating lift, propulsion, and control forces, can perform maneuvers beyond the reach of conventional fixed-wing or rotary-wing aircrafts \cite{Shyy1999FlpOverview, Shyy2010FWaeroReview, Bayiz2018HoverEff}. Compared to insect-inspired FMAVs, which excel in agility and hovering, bird-inspired FMAVs (commonly referred to as ornithopters) offer superior efficiency and robustness for sustained, long-distance flights. However, achieving robust, agile, and precise trajectory tracking control for bird-inspired FMAVs poses significant challenges. In particular, several key obstacles across various aspects of bird-inspired FMAVs design and operation hinder the development and performance of effective control systems for these vehicles.

\begin{figure}[t]
    \centering
    \includegraphics[width=0.7\linewidth]{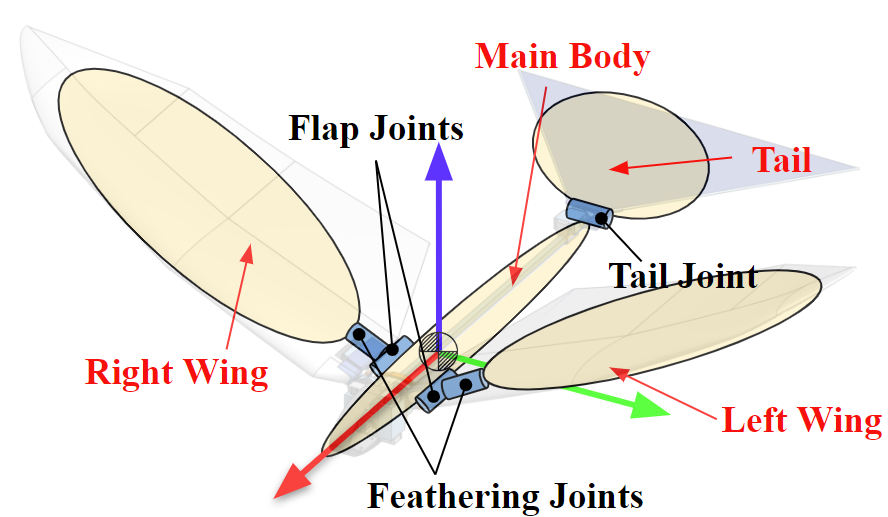}
    \caption{The layout of the flapping-wing robot. In simulation, the flapping wing robot is modeled as 4 rigid ellipsoid bodies (yellow ellipses) with 5 joints (blue cylinders). The bold red, green, and blue arrows represent the local xyz-frame of the vehicle.}
    \label{fig:flappy-fig}
    \vspace{-10pt}
\end{figure}

\subsection{Challenges for Control of Bird-inspired FMAVs}
\textbf{Modeling.} The bird-inspired FMAVs exhibit highly nonlinear aerodynamics \cite{Shyy2010FWaeroReview}. Birds and bird-inspired FMAVs typically operate within a Reynolds number range of $10^3$ to $10^6$, where the boundary layer is prone to laminar-turbulent transitions, significantly affecting the aerodynamic characteristics of flapping flight. These flow transitions are difficult to model accurately\cite{Taha2012Ctrl}. Additionally, unlike insect-scale or hummingbird-scale hoverable FMAVs (where wing inertia is often negligible), the larger wings of bird-inspired FMAVs introduce significant inertia effects. This necessitates the application of a multi-body dynamics model to capture the full complexity of the system’s movements and interactions \cite{Khosravi2019MuiltRigidBo}. Furthermore, many of the current bird-inspired FMAVs feature flexible wings that couple with the aerodynamic forces acting on them \cite{Xue2024Review}. The deformation of these wings during flight plays a critical role in determining their aerodynamic properties. This aeroelastic coupling, where aerodynamic forces and structural deformation interact, further complicates the modeling of FMAVs and presents a substantial challenge to the development of effective control strategies\cite{Taha2012Ctrl}.

\textbf{Simulation.} Given the current limitation on existing models in flapping-wing flight dynamics, aerodynamics, and aeroelastics, simulating the full flight dynamics is challenging. High-fidelity simulations are computationally intensive. For instance, Computational Fluid Dynamics - Computational Structure Dynamics (CFD-CSD) simulations usually take days to run. While the dynamics simulations for lower-fidelity models including Unsteady Vortex Latex Method (UVLM) and Unsteady Lifting Line Method are computationally feasible, they fail to account for viscous flow effects like leading-edge-vortex (LEV) and have a limited application range including the requirement of large wing aspect ratio and relatively small angle of attack for accurate results \cite{Murua2012UVLM,Boutet2018ULLT, Urban2022PteraSoftware, Sihite2022ULLT}. The lack of high-accuracy and low-computing-cost dynamical simulation environments for FMAVs is another challenge impeding the development of effective controllers for bird-inspired FMAVs. 

\textbf{Design.} From the perspective of design, the limited weight budget arising from the nature of an aerial vehicle constrains the in-flight onboard computational power. This, in turn, limits the complexity of the controller despite the complex dynamics of the vehicle. In addition, the weight constraint also restricts the number of effective actuators in the design. Most of the recent bird-inspired FMAVs are driven by a link mechanism with a single motor with gearbox for flapping, and controlling the vehicle by changing the motor speed and the angle of an actuated tail \cite{Kim2018DesignFA}. The limited control inputs restrict the agility and possible maneuvers of the FMAVs.

\begin{figure*}[t]
    \centering
    \includegraphics[width=0.8\linewidth]{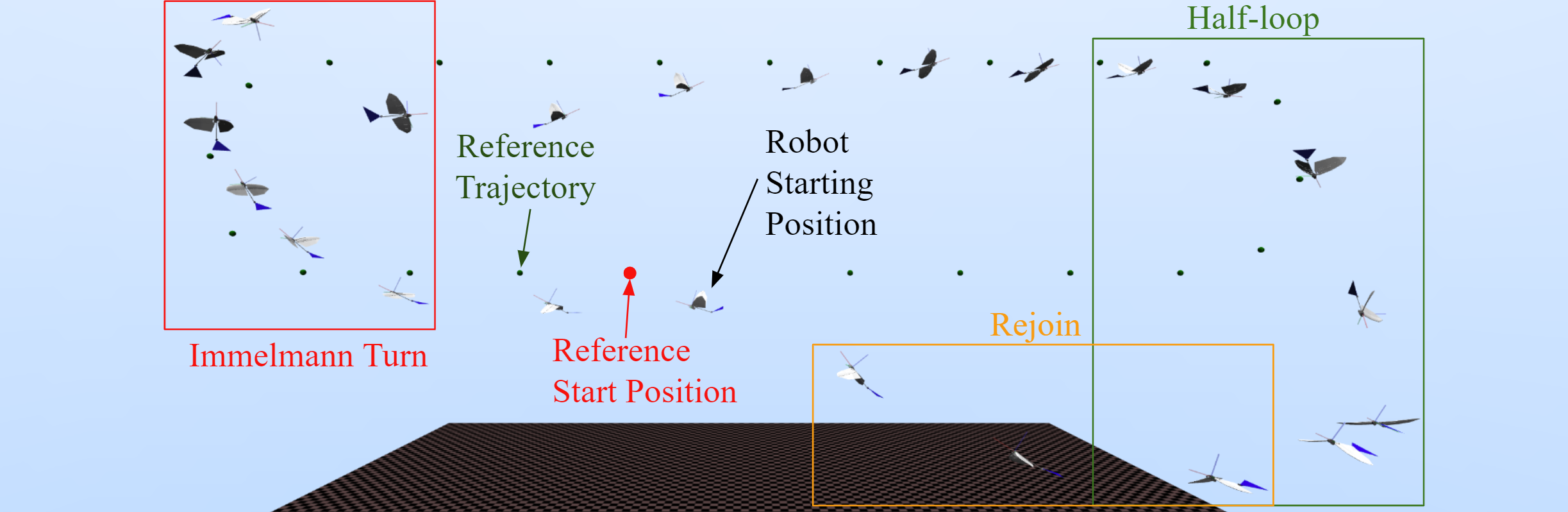}
    \caption{The flapping-wing robot follows a loop trajectory generated from simulation. The robot performs an Immelmann turn (pitch up and roll back level), a half-loop maneuver (pitch down and roll back level), and a rejoin to the trajectory. The dark green points represent the reference points of the trajectory over time.}
\label{fig:overlay_image}
\end{figure*}

\subsection{Related Work}
FMAVs can be broadly categorized into two types: insect- or hummingbird-inspired FMAVs, which are optimized for near-hover flight, and bird-inspired FMAVs, which are primarily designed for forward flight \cite{BINABAS2016FMAVReview}. Insect-sized FMAVs, due to their small size and lightweight structures, are typically more focused on hovering capabilities and agile maneuvers, often controlled with high-frequency wing flapping. In contrast, bird-inspired FMAVs are larger and more suitable for efficient forward flight, with the added complexity of managing the transition between various flight modes \cite{Liang2020ADRCland}.

Traditional control approaches for these bird-inspired flapping-wing robots often rely on model-based methods with single-rigid-body dynamics, which are generally successful in limited environments but struggle with the nonlinear, highly dynamic aerodynamic forces and complex multi-degree-of-freedom movements inherent to flapping-wing flight \cite{Xue2024Review}. These approaches include low-level attitude control systems designed for ornithopters, where proportional-integral-derivative (PID) techniques, and state feedback linearization are used to manage basic flight stabilization and orientation \cite{Torres2016OrinthCtrl,Xue2024Review}.

Some progress has been made in optimization-based approaches for flight control of bird-sized FMAVs, as demonstrated in several studies \cite{Gubta2024banking, GuptaBoundingFlight, Zhu2018MPCflap, Sihite2020CDCnonholo}. Researchers have also explored model-based methods for trajectory tracking and generation, particularly in forward flight \cite{Qian2024TrajGenTrack, Ndoye2023VFATrajTrack, He2020ModelandTraj, Hoff2019TrajBat, Li2024PreCtrlTraj}. However, these systems usually result in limited accuracy due to the simplifications required to handle the complex dynamics of flapping-wing flight. Furthermore, the control mechanisms in such systems typically involve a small number of actuators, primarily focusing on the flapping frequency and the tail’s pitch or yaw angles, which results in limited agility and maneuverability during flight. 

In contrast, learning-based approaches have recently emerged to overcome the limitations of traditional control methods. Reinforcement learning (RL) has been successfully applied to various motile platforms, including ground vehicles, drones, and legged robots, enabling robust and agile locomotion in highly dynamic environments \cite{Zhang2022Race, Song2023SR_droneRacing, Zhang2022robnav, li2024reinforcement, Zhang2024proxfly}. However, its application in flapping-wing robots remains relatively underexplored. Some early studies have applied RL for specific tasks such as improving lift generation in butterfly-like MAVs or suppressing wing vibrations during wing trajectory tracking \cite{He2019ItrLearnControl, Xiong2023ButterflyRL}. Experimental work has also demonstrated the potential for RL in enhancing efficiency in lift generation and in modeling dynamic behavior in traditional control frameworks \cite{Bayiz2019ExpLearnFlp, Lee2018LearnPathCtrl}. RL-based control has particularly shown promise in enabling agile maneuvers, such as back-flips and escape behaviors, in insect-sized hoverable FMAVs \cite{Song2010Hum_RL, Fei2019_humRL, Tu2021_FlipRL, Vaxenburg2024_fruitfly, hong2021controlflymimickingflyercomplex}. Despite these advances, the use of RL in bird-sized FMAVs for forward flight remains an open area of research, with significant potential to unlock new levels of agility and performance.

\subsection{Contribution}
This paper presents a learning-based approach for trajectory tracking in flapping-wing robots using reinforcement learning. Our method leverages a simulation environment built in MuJoCo (Multi-Joint dynamics with Contact) \cite{todorov2012mujoco} to model the robot's dynamics and aerodynamics, enabling us to train the RL policy under various conditions. We focus on developing a control framework that not only tracks predefined trajectories with high precision but also handles challenging maneuvers such as loops and Immelmann turns, as shown in Fig. \ref{fig:overlay_image}. Furthermore, we explore the robustness of the proposed method in the presence of wind disturbances and varying aerodynamic conditions, demonstrating its versatility in real-world scenarios. In addition to introducing a novel control strategy, we demonstrate the stability of the controlled system. A phase study of the system is also conducted to reveal the periodicity of the controlled system. 

\section{Problem Description}
\subsection{Platform Overview}
In this paper, we use a flapping-wing robot with a wingspan of $0.995$m, a standard mean chord of $0.17$m, and a weight of $0.31$kg, as shown in Fig.\ref{fig:flappy-fig} (a). A simplified model of the flapping-wing robot showing the joint and rigid bodies is shown in Fig. \ref{fig:flappy-fig}. Unlike many existing bird-inspired FMAVs, this robot features $ n_j = 5$  controllable joints ($\mathbf{q}_{j} \in \mathbb{R}^{5}$) that independently actuate the wing and tail. Each wing is equipped with a flap joint ($q_{1}$ for the left, $q_{3}$ for the right) to control the flap angle, and a feathering joint ($q_{2}$ for the left, $q_{4}$ for the right) to control the wing pitch. Additionally, there is a dedicated joint $q_{5}$ to control the tail pitch. In total, the robot possesses $5$ actuated DoFs. The robot also has a floating base $\mathbf{q}_{b}$ with $6$ DoFs comprising translational positions $[q_{x}, q_{y}, q_{z}]^T$ and rotational positions $[q_{\phi}, q_{\theta}, q_{\psi}]^T$, which result in a total of $n = 11$ DoFs. Thus, the system has $n-n_j = 6$ degrees of underactuation. The full system's generalized coordinates $\mathbf{q}$ can be represented as $\mathbf{q} = [\mathbf{q}_{b}, \mathbf{q}_{j}]^T \in \mathbb{R}^{n}$.

\subsection{Dynamical Model}
\label{sec:dynamics}
Although the wings of a flapping-wing robot can be flexible, in this work, we assume all bodies in the robot are rigid for simplicity. The aerodynamic effect caused by the deformation of wings is handled by domain randomization mentioned in Table \ref{table:randRange}. The full-order dynamical model of the flapping-wing robot system is derived with the Newton-Euler formulation:
\begin{equation}
    \mathbf{M}(\mathbf{q})\ddot{\mathbf{q}} + \mathbf{C}(\mathbf{q}, \dot{\mathbf{q}})\dot{\mathbf{q}} + \mathbf{G}(\mathbf{q}) =  
    \begin{bmatrix}
        \mathbf{0}_{6}\\
        \bm{\tau} 
    \end{bmatrix}
    + \mathbf{u}_\text{aero}(\mathbf{q},\mathbf{\dot q})
\end{equation}
where $\mathbf{M}(\mathbf{q})\in \mathbb{R}^{n \times n}$ is the mass matrix, $\mathbf{C}(\mathbf{q}, \dot{\mathbf{q}})\in \mathbb{R}^{n \times n}$ represents the Coriolis and centrifugal forces, and $\mathbf{G}(\mathbf{q})\in \mathbb{R}^{n}$ denotes the gravitational forces, $\bm{\tau}\in \mathbb{R}^{n_j}$ is the input joint torque, while $\mathbf{u}_{\text{aero}}(\mathbf{q},\mathbf{\dot q})\in \mathbb{R}^n$ represents the generalized aerodynamic forces.
Note that the generalized aerodynamic force, $\mathbf{u}_{\text{aero}}$, can be calculated as
 \begin{equation}
     \mathbf{u}_{\text{aero}}(\mathbf{q},\mathbf{\dot q}) = \sum_{i=1}^{m} \mathbf{J}^T_i(\mathbf{q}) \bm{F}_{\text{aero},i}(\mathbf{q},\mathbf{\dot q}),
 \end{equation}
where $\mathbf{J}_i^T(\mathbf{q}) \in \mathbb{R}^{n\times6}$ is the Jacobian transpose that maps Cartesian aerodynamics wrench $\bm{F}_{\text{aero},i}(\mathbf{q},\mathbf{\dot q}) \in \mathbb{R}^6$ acting on each body, $i$, to the generalized coordinates, and  $m$ is the number of bodies considered in the fluid. 

\subsection{Simulation}
MuJoCo is used as the physics simulator in this work. In MuJoCo, the multi-body dynamics model mentioned in section \ref{sec:dynamics} is computed. MuJoCo also provides state-less fluid force models (i.e., the fluid force computation does not have its own dynamics and fluid states but rather is only dependent on the state of the robot - a simplifying approximation that is made), which is used to model the Cartesian aerodynamic wrench $\bm{F}_{\text{aero}}=[\mathbf{f}^T_{\text{aero}}~\boldsymbol{\tau}^T_{\text{aero}}]^T$ acting on each body $i$ of the robot. There are two different models provided by MuJoCo that were both used in this work: (1) a simplified inertia model that estimates the aerodynamic wrench based on its equivalent inertia box, and (2) a more elaborate ellipsoid model that accounts for the aerodynamic force and moment from 5 different aerodynamic effects on a projected ellipsoid \cite{Vaxenburg2024_fruitfly}.
For a single rigid body in the fluid, the total aerodynamic force, $\mathbf{f}_\text{aero}$, and moment $\boldsymbol{\tau}_\text{aero}$ are calculated as follows for each model:
For ellipsoid model, 
\begin{align}
\mathbf{f}_\text{aero} &= \mathbf{f}_A + \mathbf{f}_D + \mathbf{f}_M + \mathbf{f}_K + \mathbf{f}_V \\
\boldsymbol{\tau}_\text{aero} &= \boldsymbol{\tau}_A + \boldsymbol{\tau}_D +\boldsymbol{\tau}_V
\end{align}

and for inertia model, 

\begin{align}
    \mathbf{f}_\text{aero} &= \mathbf{f}_A + \mathbf{f}_V \\
    \boldsymbol{\tau}_\text{aero} &=\boldsymbol{\tau}_A + \boldsymbol{\tau}_V
\end{align}

where subscripts $A$, $D$, $M$, $K$, and $V$ represent added mass, viscous drag, Magnus lift, Kutta lift, and viscous resistance, respectively. The computation of each term is summarized as follows based on \cite{Vaxenburg2024_fruitfly},

\begin{equation}
\begin{aligned}
\mathbf{f}_A &= -\mathbf{m}_A \circ \dot{\mathbf{v}} + (\mathbf{m}_A \circ \mathbf{v}) \times \boldsymbol{\omega}\\
\boldsymbol{\tau}_A &= -\mathbf{I}_A \circ \dot{\boldsymbol{\omega}} + (\mathbf{m}_A \circ \mathbf{v}) \times \mathbf{v} + (\mathbf{I}_A \circ \boldsymbol{\omega}) \times \boldsymbol{\omega}\\
\mathbf{f}_D &= -\rho \left[C_{D, \text{blunt}} A_{\text{proj}}^{\mathbf{v}} + C_{D, \text{slender}} (A_{\max} - A_{\text{proj}}^{\mathbf{v}})\right] \lVert \mathbf{v} \rVert \mathbf{v}\\
\boldsymbol{\tau}_D &= -\rho \left[C_{D, \text{angular}} I_D + C_{D, \text{slender}} (I_{\max} - I_D)\right] \boldsymbol{\omega}\\
\mathbf{f}_M &= C_M \rho V \boldsymbol{\omega} \times \mathbf{v}\\
\mathbf{f}_K &= C_K \rho A_{\text{proj}}^{\mathbf{v}} \lVert \mathbf{v} \rVert (\mathbf{v} \times \mathbf{v}_{\parallel}) \times \mathbf{v}\\
\mathbf{f}_V &=-6 \pi r_V \nu \mathbf{v} \\
\boldsymbol{\tau}_V &=-8 \pi r_V^3 \nu \boldsymbol{\omega}
\end{aligned}
\end{equation}

\begin{table}[t]
\centering
\caption{Definition of Variables Used in the Aerodynamics Equations}
\begin{tabular}{|c|m{5cm}|}
\hline
\textbf{Variable} & \textbf{Definition} \\
\hline
$\mathbf{v}$ & Velocity (vector) relative to airflow \\ \hline 
$\mathbf{v}_{\parallel}$ & Component of velocity parallel to surface \\
\hline
$\boldsymbol{\omega}$ & Angular velocity (vector) \\ \hline 
$\rho$ & Fluid density \\ \hline 
$\nu$ & Kinematic viscosity of the fluid \\ \hline
$\mathbf{m}_A$ & Added mass (vector) \\ \hline 
$\mathbf{I}_A$ & Added moment of inertia (vector) \\ \hline 
$C_{\text{D, blunt}}$ & Drag coefficient for blunt body \\ \hline 
$C_{\text{D, slender}}$ & Drag coefficient for slender body \\ \hline 
$C_{\text{D, angular}}$ & Angular drag coefficient \\ \hline 
$C_K$ & Kutta lift coefficient \\ \hline 
$C_M$ & Magnus force coefficient \\ \hline 
$A_{\mathbf{v}}^{\text{proj}}$& Projected area in the direction of velocity \\ \hline 
$A_{\max}$ & Maximum reference area \\ \hline 
$V$ & Volume of the body\\ \hline 
$A_{\max}$ & Maximum reference area \\ \hline 
$I_D$ & Reference moment of inertia for drag\\ \hline
$r_V$ & Effective radius for viscous drag \\ \hline 
\end{tabular}
\label{table:aeroTerms}
\end{table}
where $\circ$ is defined as element-wise multiplication, and each of the terms used is defined in Table \ref{table:aeroTerms}. Note that in this work, the fluid coefficients $C_{D, \text{blunt}}$, $C_{D, \text{slender}}$, $C_{D, \text{angular}}$, $C_K$, and $C_M$ are manually tuned to match the designed lift-to-drag ratio of the FMAV at gliding, given that the horizontal-to-vertical distance covered by the FMAV is equal to the lift-to-drag ratio when gliding in steady state. The resultant coefficients are provided in Table \ref{table:FluidCoef}. For higher simulation accuracy, we model all the lifting bodies including wings and tail with the more elaborate ellipsoid model and model the remaining bodies, i.e., the main body, with inertia model in considering the relatively small aerodynamic effect from the body compared to the lifting bodies. This enables us to capture the most significant aerodynamic properties of the robot in MuJoCo with a relatively simple model.
\begin{table}[t]
    \centering
    \caption{Aerodynamic coefficients used in this work.}
    \begin{tabular}{|c|c|c|c|c|c|} \hline 
         Coefficient&  $C_{\text{D, blunt}}$ &  $C_{\text{D, slender}}$ &  $C_{\text{D, angular}}$ & $C_K$  & $C_M$\\ \hline 
         Value&  0.2 &  0.12 &  1.5 &  3.14 &  1\\ \hline
    \end{tabular}
\label{table:FluidCoef}
\end{table}

\section{RL-Based Trajectory Tracking Control}

\subsection{Control Framework}
\begin{figure}[t]
    \centering
    \includegraphics[width=\linewidth]{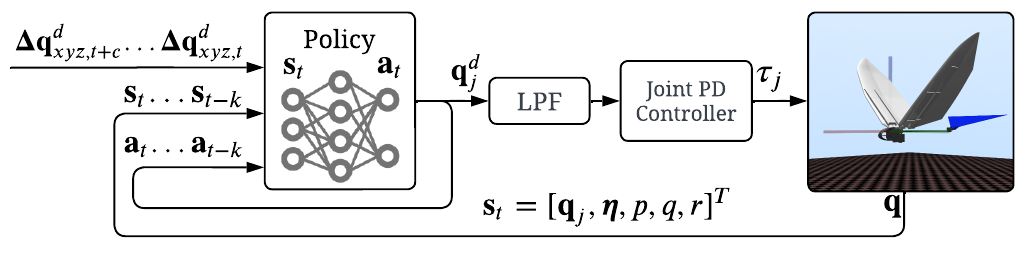}
    \caption{The control diagram for flapping-wing robot trajectory tracking control. The variables used here are covered in Sec \ref{sec:StateAct_Spaces}}.
    \label{fig:ctrl-framework}
    \vspace{-10pt}
\end{figure}

Our goal is to develop an RL controller framework for a bird-inspired FMAV that can track target trajectories and achieve bird-like maneuverability. As seen in Figure~\ref{fig:ctrl-framework}, we use an RL-trained policy that outputs actions that are scaled and smoothed via a low-pass filter into target joint positions. These are then used by the low-level PD controller to compute motor torques. The policy operates at 50Hz, while the low-level joint PD controller runs at the simulation frequency of 250Hz. \\
One important aspect of our framework was ensuring energy-efficient behavior, since policies tend to prefer unrealistically high flapping frequencies when left unconstrained. Aside from the enforced energy penalties during training, the low-pass filter was given a 7Hz cutoff frequency to limit the viable flapping frequencies to 4-6Hz.\\
Due to the difficulty of adapting to a variety of arbitrary maneuvers as well as lack of a ``safe state" in our state space, we found it difficult to train such a policy from scratch. Therefore, we use a curriculum-based training scheme involving 3 stages of increasing difficulty to train a controller: (1) constant forward flight, (2) climbing and diving at variable speeds, (3) turning and arbitrary maneuvers. Following this, Dynamics randomization was an additional stage to improve the robustness and adaptability of trained controllers, where we randomized rigid body, motor-level characteristics, the aerodynamics and wind condition of the system. 

\subsection{Target Trajectories}
\label{sec:Target_Trajectories}
We procedurally generate 3D position trajectories $\mathbf{Q}^{d} _{xyz}(t)$ that are defined by simple linear and circular paths. The policy is provided with a look-ahead buffer of the upcoming trajectory and is rewarded based on its position error to the current target position along it. These simple trajectories were chosen over trajectories obtained through model-based optimization and Bézier curves \cite{A1ShootingJi2022} due to uncertainty in aerodynamic feasibility. The robot's flight path is not tightly constrained to this target trajectory to allow the policy to optimize its movement according to the aerodynamics of the system. Hence, this ensures that the learning is unconstrained and natural, with the resulting emergent behavior dictating its own orientation to fit the desired movement. With the parameters of forward speed, Z-axis velocity, and global angular yaw rate, we can specify commands that combine the four basic flying skills: flying straight, diving, climbing, and turning. The duration of each command is fixed at 3 seconds, enabling the policy to follow arbitrary paths and become robust to skill transitions. Vertical loops were also modeled to simulate aerobatic maneuvers like back-flips and Immelmann turns.

\subsection{State and Action Spaces}
\label{sec:StateAct_Spaces}
The actions $\mathbf{a}_{t} \in \mathbb{R}^{n_{q}}$ that are output from the controller specify the target actuated joint positions $\mathbf{q}_{j}^d$, which are used by the low-level joint PD controller to compute the motor torque. The action space is centered in the nominal pose (wings and tail flat at 0 degrees) and is normalized to the respective joint limits. 

The observation space of the policy consists of sensor-related observations and trajectory information. The sensor observations for each time frame include the orientation quaternion $\bm{\eta}$, local angular velocity $p,q,r$, joint motor positions $\mathbf{q}_{j}$, and a local x-velocity measurement relative to the wind $\mathbf{v}_{x,\text{air}}$. The latter sensor is modeled after a pitot tube and acts as the most plausible form for a physical flapping robot to receive onboard velocity readings. The policy receives a history of sensor readings and past action outputs for the past 25 steps, corresponding to a window of $0.5$s. This history allows the policy to infer the dynamics of the system.

The policy also receives 30 steps of the upcoming trajectory $\mathbf{Q}_{xyz} ^{d} (t)$ in local-frame relative coordinates $\Delta \mathbf{q}_{xyz} ^{d}$. This corresponds to $0.6\text{s}$ of the future trajectory information. Although shorter trajectory windows of $0.2\text{s}$ have also resulted in well-performing policies, larger windows are included to allow for smooth flapping behavior.

\subsection{Rewards}
The heuristics for our reward $r$ for the policy are kept constant across all stages of training: to minimize position error relative to the target trajectory while preserving balance. The reward function has four parts
\begin{equation}
    r = 0.5 r_{\text{pos}} + 0.1r_{\mathbf{\Omega}} + 0.2 r_{\phi, \theta} + 0.05 r_{\text{energy}}
\end{equation}
where $r_{\text{pos}}$ is the position tracking term that minimizes body position error to the current desired position on the target trajectory. $r_{\mathbf{\Omega}}$ attempts to minimize the main body angular rates $p, q, r$ to ensure smoother and stable behavior. $r_{\phi, \theta}$ rewards a lower roll and pitch, motivating the robot to stay leveled. Although a bird's pitch and roll are involved in its flight movements, this reward ensures that the learned behavior varies its orientation only when needed. This is because a bird receives maximal horizontal thrust when oriented horizontally, and hence this reward encourages the policy to maximize its time in this orientation.  $r_{\text{energy}}$ is the energy term \cite{fu2021minimizing} that motivates energy-efficient behavior, i.e., gliding instead of flapping when possible, mimicking physical birds. This term also serves the purpose of discouraging high flapping frequencies so that the learned frequency resembles those of physical birds of the same wing area.

\subsection{Dynamics and Aerodynamic Force Randomization}
\label{sec:rand}

We have incorporated the randomization of dynamic parameters in our training framework in order to better prepare policies for the sim-to-real gap. Similar to previous work \cite{A1ShootingJi2022}, randomization was applied to rigid body mass, inertia, and center of mass.

Although the Mujoco Fluid model is capable of capturing salient aerodynamic effects using ellipsoids, it still has limited accuracy to simulate the real physics of the flapping-wing robot. This is due to three major reasons: (1) the hand-tuned fluid coefficients may not match the physical model, (2) it's difficult for the state-less characteristic of MoJoCo aerodynamic model to capture the state-dependent unsteady aerodynamic force in highly dynamical motions, and (3) the deformation on the physical wing can cause variations in the resultant aerodynamic forces. To overcome this deviation between the simulation and the real world, a randomization on the aerodynamics is incorporated. This includes randomizing the 5 different fluid coefficients, the added mass $\mathbf{m}_A$, and the added inertia $\mathbf{I}_A$ mentioned in Table \ref{table:aeroTerms} on each body. Additional wind disturbances with randomized direction and magnitude were also implemented to improve the robustness and adaptability of the policy. The detailed range of dynamics randomization is shown in Table \ref{table:randRange}. The initial position and velocity are also randomized to ensure that the policy can robustly recover
and rejoin trajectories.
\begin{table}[htbp]
\centering
\caption{The range of dynamics randomization. }
\label{table:randRange}
\begin{tabular}{ll}
\toprule
\textbf{Parameters} & \textbf{Range} \\
\midrule
Joint Damping Ratio & [0.9, 1.1] Nms/rad \\
Link Mass \& Link Inertia & [0.9, 1.1] $\times$ default \\
Link CoM Position & [-0.05, 0.05] m + default \\
Aerodynamic Coefficients  & [0.7, 1.3] $\times$ default \\
Aerodynamic Added Mass \&  Inertia   & [0.9, 1.1] $\times$ default\\
Wind x,y,z Velocity & [$\pm$2], [$\pm$2], [$\pm$1.5]  m/s\\
\bottomrule
\end{tabular}
\end{table}

\subsection{Episode Design}
Each episode lasts a maximum of $\SI{30}{\second}$, corresponding to 1,500 control steps. A position error termination condition is enforced throughout all stages of control, terminating episodes if the robot deviates more than 3 meters from the target position. This condition accelerates early training and improves the precision of trajectory tracking while allowing the policy to optimize its own flight path. Additionally, an orientation termination condition is also applied during the first two stages of training to ensure that the robot's absolute roll $|q_{\phi}|$ and pitch $|q_{\theta}|$ never exceeded $90^\circ$ when learning basic flight. This constraint is removed when training later stages that involve arbitrary paths and extreme maneuvers. We trained our policy with a multilayer perceptron (MLP) actor-critic architecture with Proximal Policy Optimization (PPO) \cite{schulman2017proximal} employing 256 parallel environments. Each training stage required approximately 8 million steps, with a full controller trained in about 30 million steps. The training was conducted in CPU with an Intel Core i7-11800H and took around 5 hours to train the full controller.

\section{Results}
\begin{figure}[t]
    \centering
    \includegraphics[trim = {0 1cm 0 0}, clip, width=\linewidth]{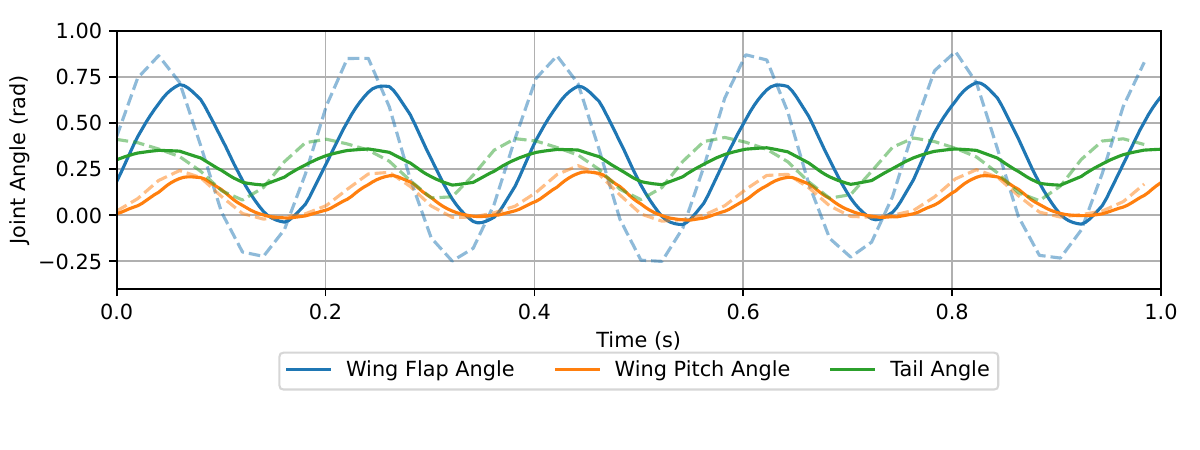}
    \caption{Forward flapping behavior of the controller shown through time series of the wing pitch angle, wing pitch angle, and tail angle in 1 second. The dashed lines are target joint positions while the solid lines indicate the actual joint positions.}
\label{fig:forward-success}
\end{figure}

We now validate the performance of policies trained for our control framework. Fig. \ref{fig:forward-success} displays 1 second of flapping at a speed of $\SI{3.8}{\meter\per\second}$. The wing flapping, wing pitch, and tail angles were found to have a fundamental frequency of 5.3Hz with energies of 38.65\%, 36.81\%, and 38.24\% respectively, indicating reasonable fit to a single frequency mode. 

\subsection{Stability Analysis of the Flight Controller}
\subsubsection{System Identification}
\begin{figure}
    \centering
    \includegraphics[trim = {0 0 0 0.2cm}, clip, width=8.6cm]{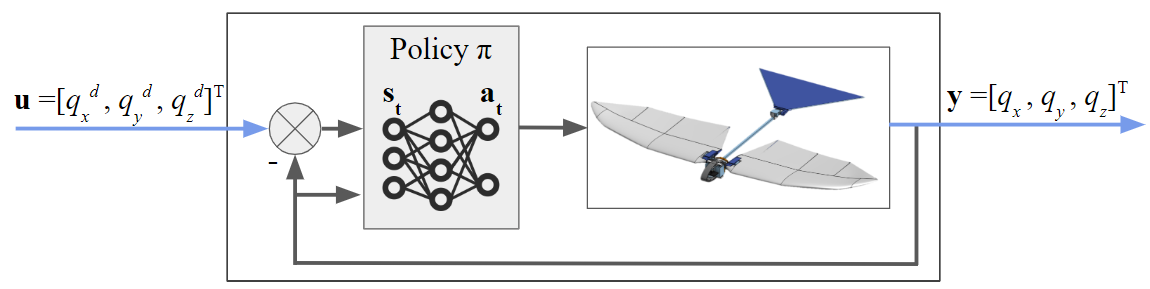}
    \caption{System identification is performed on the closed-loop system. A low-dimensional system is derived from the high-dimensional, nonlinear dynamics of the flapping-wing robot, which is controlled by its RL policy. The input, $\mathbf{u}$, of this simplified system is the desired global position, while the output, $\mathbf{y}$, represents the robot’s measured response, driven by the low-level RL policy.}
    \label{fig:diag_LowDimSys}
\end{figure}

To analyze the dynamics of the RL-based controller and validate the stability of the control system, a model derived from system identification using input-output pairs is introduced, as shown in Fig. \ref{fig:diag_LowDimSys}. Following a method similar to \cite{li2022bridging}, a low-dimensional linear system is extracted from the flapping-wing robot under the control of the RL-based controller, which is then employed to demonstrate its stability. 
The input $u$ of the closed-loop system is the desired position $[q_{x}^d, q_{y}^d, q_{z}^d]^T \in \mathbb{R}^3$ while the output of the system is $\mathbf{y} = [\hat{q_{x}}, \hat{q_{y}}, \hat{q_{z}}]^T$. We develop a linear model that approximates the behavior of the closed-loop system governed by the model-free RL-based policy. The LTI system is obtained by fitting the input-output pairs, where the input of the closed system is determined by the input of the policy network.
The fitted input-output dynamics of position is given by:

\begin{align}
    Y_{x}(s) &= \frac{49.89s^2 + 164.9s + 26.27}{s^3 + 3.554s^2 + 6.438s + 2.809}\\
    Y_{y}(s) &= \frac{-0.09798s^2 - 10.07s - 24.67}{s^3 + 3.554s^2 + 6.438s + 2.809}\\
    Y_{z}(s) &= \frac{1.006s^2 + 1.020s + 3.836}{s^3 + 3.554s^2 + 6.438s + 2.809}
\end{align}

Note that the derived linear model predicted result with a Mean Squared Error (MSE) as low as $5.629\times10^{-5}$.

\subsubsection{Stability}
\begin{figure}[t]
    \centering
    \begin{subfigure}[t]{0.32\linewidth} 
        \centering
        \includegraphics[width=\linewidth]{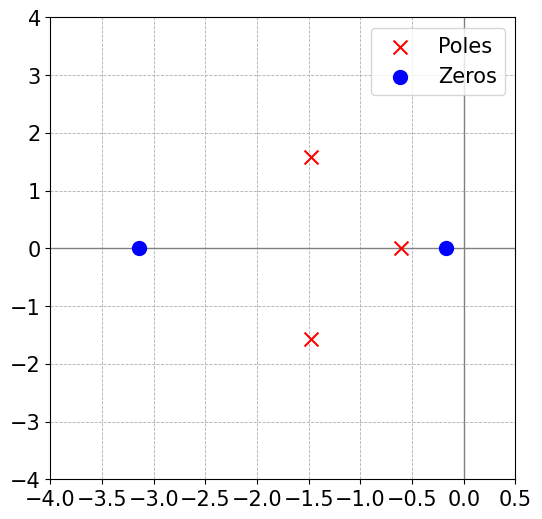}
        \caption{Pole-zero of $q_{x}$}
    \end{subfigure}
    \hfill
    \begin{subfigure}[t]{0.32\linewidth} 
        \centering
        \includegraphics[width=\linewidth]{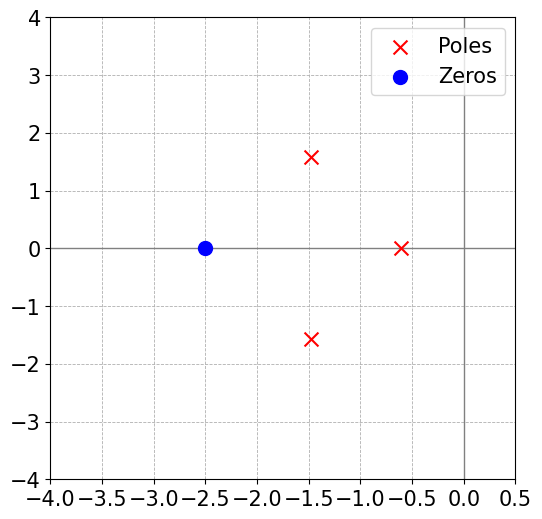}
        \caption{Pole-zero of $q_{y}$}
    \end{subfigure}
    \hfill
    \begin{subfigure}[t]{0.32\linewidth} 
        \centering
        \includegraphics[width=\linewidth]{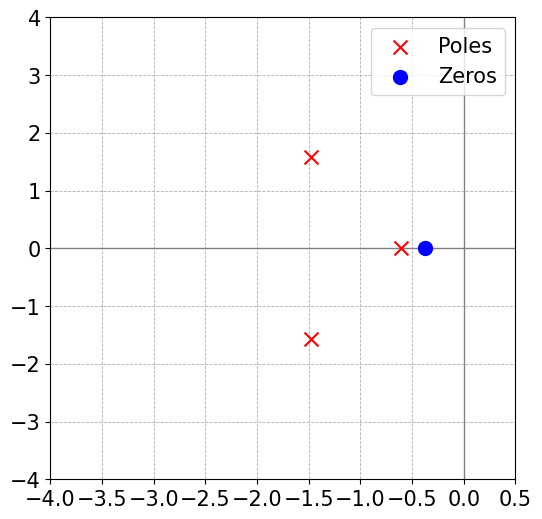}
        \caption{Pole-zero of $q_{z}$}
    \end{subfigure}    
    \caption{Pole-zero plots for the closed-loop system’s linear dynamics in three dimensions. In all three plots, the poles are located in the LHP, indicating that the system exhibits bounded input bounded output (BIBO) stability across all dimensions. Additionally, the zeros are also located in the LHP, confirming that the system is minimum phase in each dimension.}
    \label{fig:pole_zero_plots}
\end{figure}

We assess the stability of the closed-loop system's linear dynamics by examining the pole positions of the above transfer functions. The plots of poles and zeros for all three dimensions are presented in Fig. \ref{fig:pole_zero_plots}. Based on the analysis, all identified linear systems exhibit Bounded Input Bounded Output (BIBO) stability, as all poles are located in the left-half plane (LHP). This shows that the input-output dynamics of the closed-loop system comprised of the nonlinear FMAV controlled by the RL policy is locally input-output stable. Additionally, all zeros are located in the LHP, confirming that the system is minimum phase in all dimensions. The result implies that the input-output relationship of our system does not exhibit non-minimum phase behavior, which can often be found in conventional fixed-wing aircraft, where tail-down effects precede tail-up behavior. Therefore, the choice of input-output plays a critical role in achieving a minimum phase, making the closed-loop system more controllable and stabilizable.
\subsection{Phase portraits of Flying Tasks}
\begin{figure}[t]
    \centering
    \begin{subfigure}[t]{0.49\linewidth} 
        \centering
        \includegraphics[width=\linewidth]{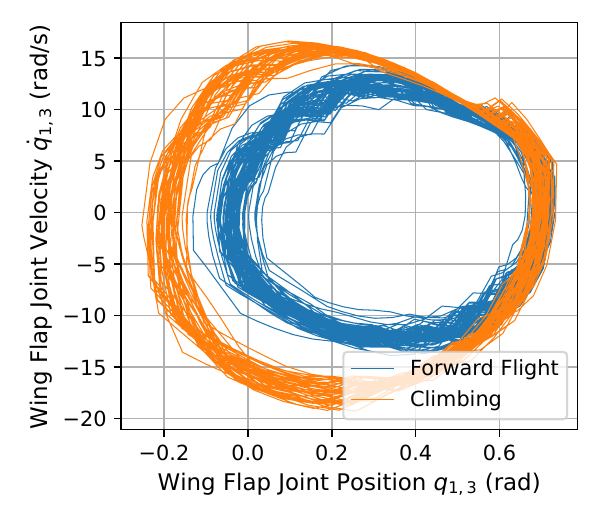}
        \caption{Wing Flap joint}
    \end{subfigure}
    \hfill
    \begin{subfigure}[t]{0.49\linewidth}
        \centering
        \includegraphics[width=\linewidth]{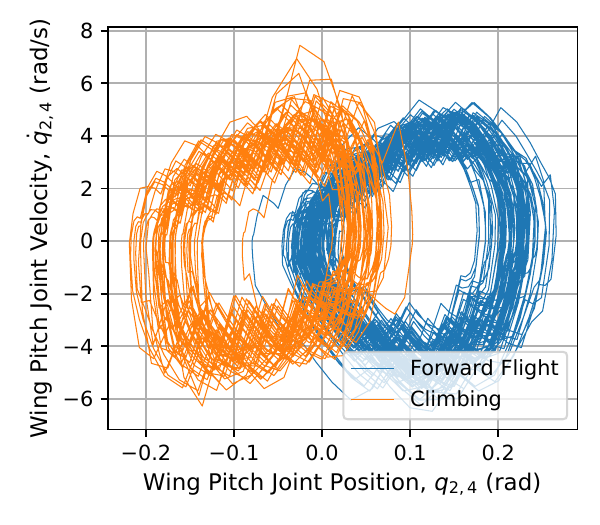}
        \caption{Wing Pitch joint}
    \end{subfigure}
    \caption{Comparing phase portraits between cruising and climbing}
    \label{fig:phase-flap-pitch_c}
\end{figure}

\begin{figure}[t]
    \centering
    \begin{subfigure}[t]{0.49\linewidth} 
        \centering
        \includegraphics[width=\linewidth]{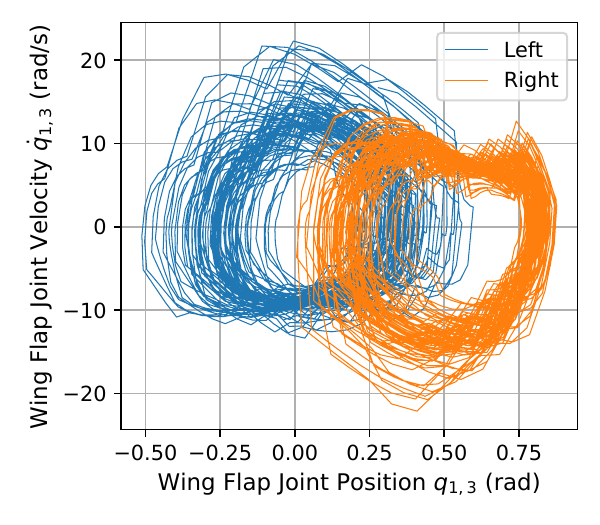}
        \caption{Wing Flap joint}
    \end{subfigure}
    \hfill
    \begin{subfigure}[t]{0.49\linewidth} 
        \centering
        \includegraphics[width=\linewidth]{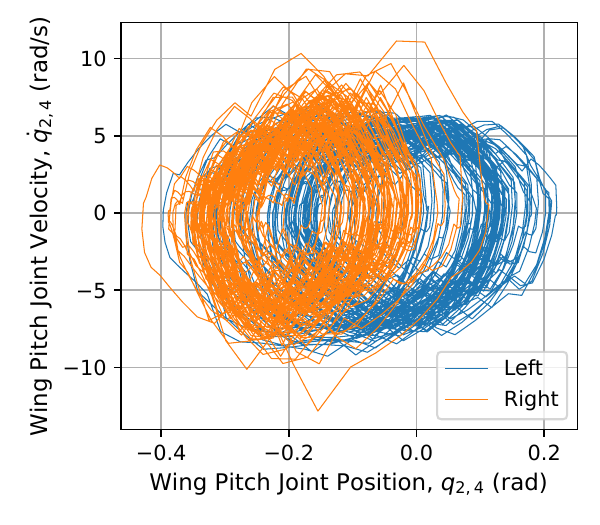}
        \caption{Wing Pitch joint}
    \end{subfigure}
    \caption{Comparing left and right wing phase portraits when turning left}
    \label{fig:phase-flap-pitch_t}
\end{figure}

To illustrate the system's periodicity under different tasks, we present phase portraits of the wing flap and wing pitch joints when undergoing forward flight, climbing, and turning. These portraits exhibit closed, consistent periodic orbits, indicating that the robot’s flapping and pitching are stable and periodic throughout flight. Fig. \ref{fig:phase-flap-pitch_c} shows wing's pitch and flap angles when flying forward and climbing at $\SI{3}{\meter\per\second}$ forward velocity. We plot the phase of only the right wing as the robot's symmetry ensures identical behavior on the left. As shown in Fig. \ref{fig:phase-flap-pitch_c} (a), climbing flight tends to have a larger range of motion for flap joint position and velocity, which is consistent with the fact that robot consumes more energy when climbing up. On the other hand, the average pitch joint angles in Fig.\ref{fig:phase-flap-pitch_c} (b) shift to negative values while retaining the same periodic motion. Note that the negative pitch angle is consistent with the body frame, where pitching up corresponds to more negative values. This implies that the policy tends to acquire a higher angle of attack to gather more lift for climbing. We then compare the flap and pitch joints between the left and right wings while gradually turning left, as seen in Fig. \ref{fig:phase-flap-pitch_t}. The right wing is raised (higher flap joint position) and pitches up (more negative pitch joint position) to generate more lift on the right. This results in higher lift and thrust generation on the right wing, thereby turning left.
\subsection{Trajectory Tracking Performance}
\begin{figure}[t]
    \centering
    \begin{subfigure}[t]{0.49\linewidth} 
        \centering
        \includegraphics[width=\linewidth]{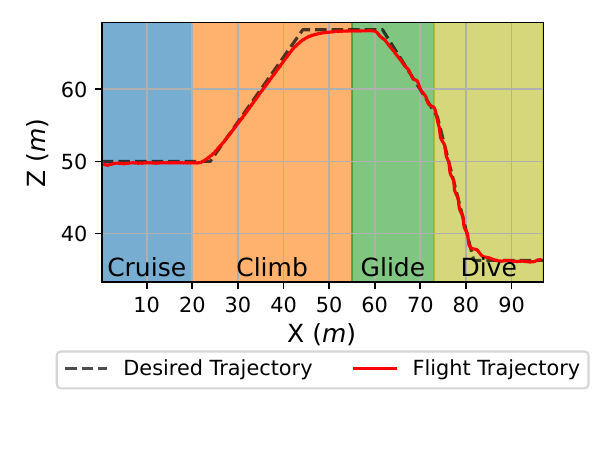}
        \caption{Longitudinal Flight Tasks}
    \end{subfigure}
    \hfill
    \begin{subfigure}[t]{0.49\linewidth} 
        \centering
        \includegraphics[width=\linewidth]{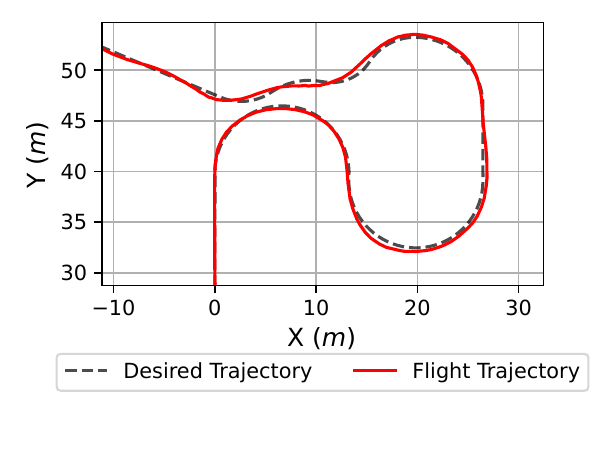}
        \caption{Lateral Flight tasks}
    \end{subfigure}
    \caption{Example longitudinal and lateral trajectories to display the controller's tracking performance. (a) A series of cruising (blue), climbing (orange), gliding (green), diving (yellow) in 24 seconds with a 3.8m/s follow velocity. (b) A series of gradual and quick turns in 18 seconds with a $\SI{5.3}{\meter\per\second}$ follow velocity.}
    \label{fig:controller_xy_xz_tracking}
\end{figure}

\begin{figure}[t]
    \centering
    \begin{subfigure}[t]{0.49\linewidth} 
        \centering
        \includegraphics[trim={0.4cm 0.7cm 0.3cm 0.3cm}, clip, width=\linewidth]{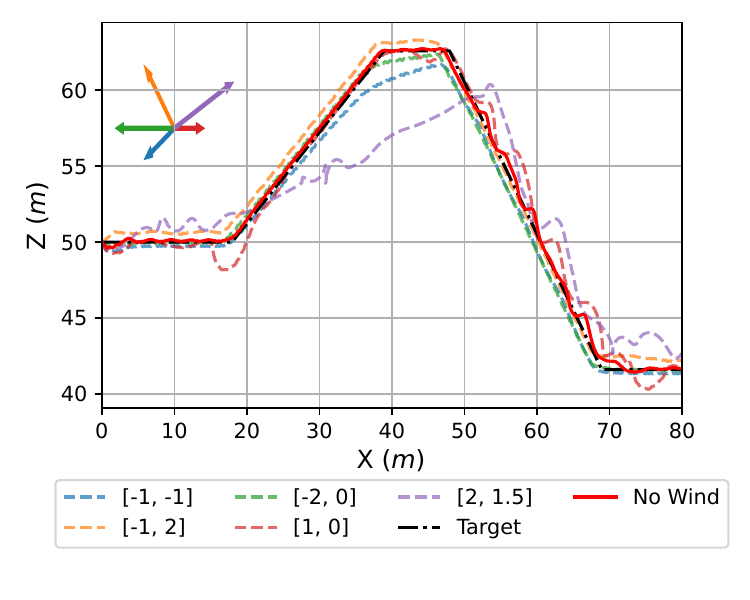}
        \caption{Simple forward flight tasks.}
    \end{subfigure}
    \hfill
    \begin{subfigure}[t]{0.49\linewidth}
        \centering
        \includegraphics[trim={0.2cm 0.5cm 0.1cm 0.3cm}, clip, width=\linewidth]{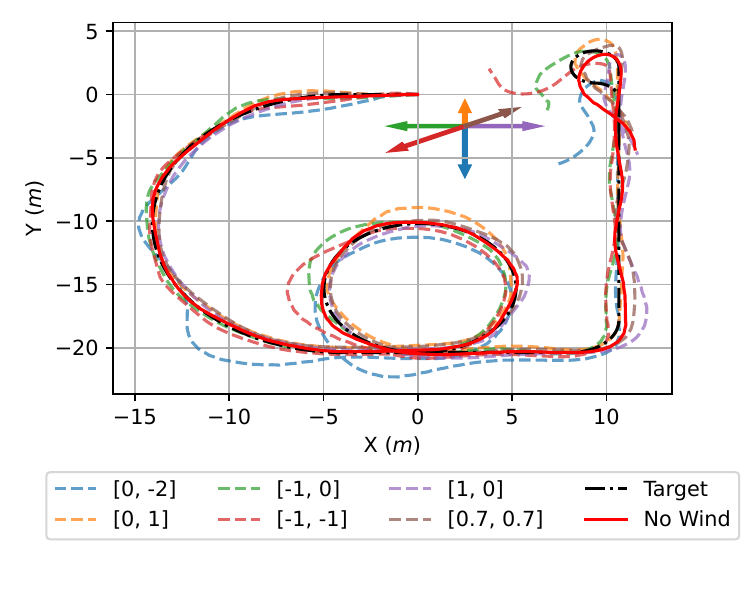}
        \caption{Turning}
    \end{subfigure}

    \caption{Effect of randomization on trajectory following. The legend shows the 2D wind vectors in $\SI{}{\meter\per\second}$ in the corresponding plane for each respective trajectory. These vectors are also visualized in the plot.}
    \label{fig:wind_xz_xy_traj}
    \vspace{-10pt}
\end{figure}

\begin{figure}[t]
    \centering
    \includegraphics[width=1\linewidth]{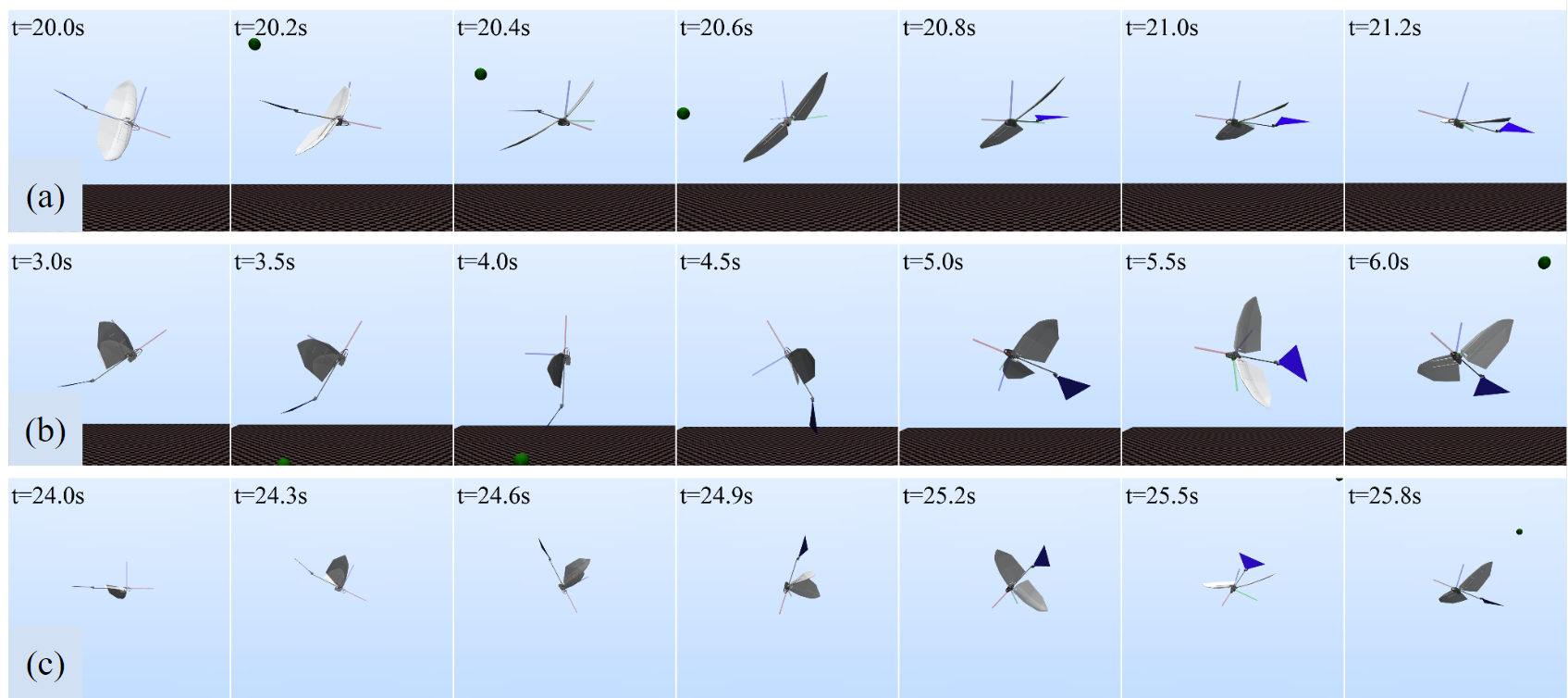}
    \caption{Snapshots of three difference aerobatic maneuvers: (a) A sharp $180^\circ$ turn maneuver in $\SI{1.2}{\second}$. (b) A loop maneuver (back-flip) over $\SI{3}{\second}$. (c) A roll-off-the-bottom maneuver (half loop pitching down followed by a roll back to level) in $\SI{1.8}{\second}$.}
    \label{fig::maneuvers}
\end{figure}

\begin{figure}[t]
    \centering
    \includegraphics[trim={1cm 0.9cm 0 1.3cm}, clip, width=0.6\linewidth]{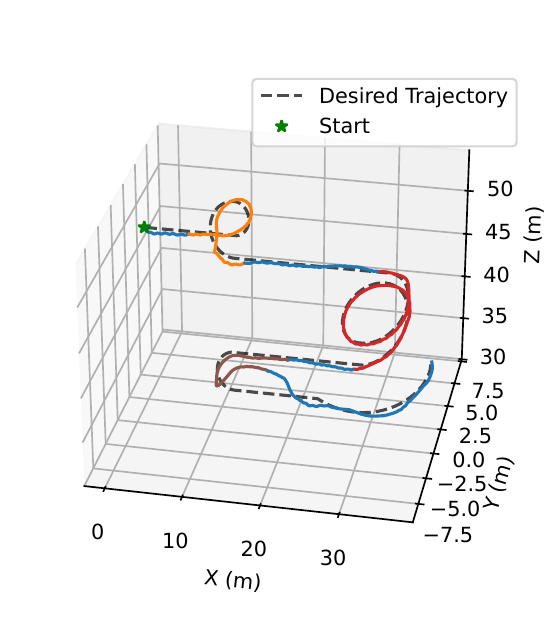}
    \caption{An aerobatic trajectory consisting of a loop (orange), turns (red), and a roll-off-the-bottom maneuver (brown). The blue sections show cruising and recovery.}
    \label{fig:AerobaticTraj}
\end{figure}

A series of trajectories were generated to validate the performance of the trained policies. Simple trajectories demonstrating fundamental flight skills like cruising, climbing, gliding, diving, and turning are shown in Fig. \ref{fig:controller_xy_xz_tracking}. The target lookahead buffer from Section \ref{sec:Target_Trajectories} allows the policies to anticipate command changes and accordingly optimize its flight path. We designed an aerobatic trajectory to demonstrate the versatility of the closed-loop RL controller on the FMAV, shown in Fig. \ref{fig:AerobaticTraj}. The corresponding snapshots of each maneuver in the same trajectory are shown in Fig. \ref{fig::maneuvers}. These results show the RL controller's capability to track highly dynamic trajectories by utilizing various combinations of control inputs to manipulate its attitude. A few other examples of flapping-wing robot tracking trajectories are also shown in Fig. \ref{fig:overlay_image} and posted at \url{https://youtu.be/54Gcbvgfz7Q}. Note that there is a noticeable tracking error when we provide a potentially dynamically infeasible target trajectory, but the FMAV is capable of rejoining the trajectory to some extent. Fig. \ref{fig:wind_xz_xy_traj} shows the controller's robustness to wind and aerodynamics randomization, illustrating each wind vector and fluid coefficient's influence on the flight path. This demonstrates that the FMAV is most sensitive to $C_K$, the Kutta lift coefficient, while still achieving a relatively high success rate even with other fluid coefficients randomized by up to a factor of 0.5. The high sensitivity to $C_K$ can be caused by its dominant role in generating lift for bird-scale flapping flight.  

\begin{figure}[t]
    \centering
    \includegraphics[width=0.6\linewidth]{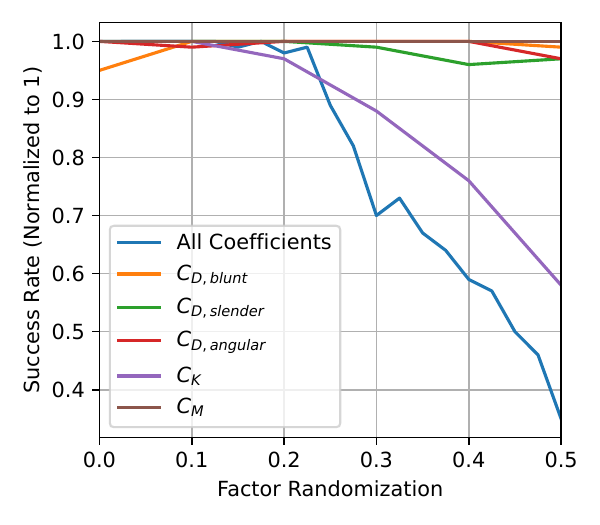}
    \caption{Randomizes specific fluid coefficients while others remain at the nominal values in Table \ref{table:FluidCoef}. The success rate is the fraction of 100 randomly sampled episodes that were within 3m at the end of the Fig. \ref{fig:wind_xz_xy_traj} (a) trajectory. All coefficients are uniformly randomized in the blue curve.}
    \label{fig::aero_sucess}
\end{figure}

\section{Conclusion and Future Work}
In this work, we proposed a novel reinforcement learning (RL)-based framework for trajectory tracking in bird-inspired flapping-wing robots. This developed control system demonstrates the ability to track complex 3D trajectories, perform agile maneuvers, and adapt to varying aerodynamic conditions in simulation. By leveraging MuJoCo's multi-body dynamic modeling and aerodynamic randomization, we ensured that the RL policy generalized well to diverse flight scenarios, including wind disturbances and different aerodynamics. Our stability analysis validated that the closed-loop system was both asymptotically stable and capable of maintaining stable, periodic joint action patterns.

It is worthwhile mentioning that the hardware platform of the flapping-wing robot is in the process of design. Our future work will involve experimental validation of the proposed RL policy once we design and build a flapping-wing robot.

\section*{Acknowledgement}
This work is inspired from discussions on dynamics and control of FMAVs with Alireza Ramezani and Bibek Gupta from Northeastern University. This work is supported in part by National Science Foundation Grant CMMI-2140650 and in part by The AI Institute. We thank Tingyu Guo and Zhongyu Li for developing preliminary RL policies related to this project. We also thank Qiayuan Liao for the technical discussions and Ruiqi Zhang for proofreading. Koushil Sreenath has financial interests in Boston Dynamics AI Institute LLC. He and the company may benefit from the commercialization of the results of this research.

\bibliographystyle{ieeetr}
\bibliography{main}

\end{document}